\documentclass[10pt,twocolumn,letterpaper]{article}

\usepackage{iccv}
\usepackage{times}
\usepackage{epsfig}
\usepackage{graphicx}
\usepackage{amsmath}
\usepackage{amssymb}
\usepackage[accsupp]{axessibility}  
\usepackage{booktabs}

\usepackage{tikz}
\usepackage{comment}

\usepackage{color}
\usepackage{multirow}
\usepackage{caption}

\usepackage{pifont}
\usepackage{mathtools}
\usepackage{algorithm}
\usepackage[noend]{algpseudocode}
\usepackage{wrapfig}

\newcommand{\cmark}{\ding{51}}
\newcommand{\xmark}{\ding{55}}

\usepackage[pagebackref=true,breaklinks=true,colorlinks,bookmarks=false]{hyperref}

\usepackage[capitalize]{cleveref}
\crefname{section}{Sec.}{Secs.}
\Crefname{section}{Section}{Sections}
\Crefname{table}{Table}{Tables}
\crefname{table}{Tab.}{Tabs.}

\iccvfinalcopy 

\def\iccvPaperID{4212} 

\ificcvfinal\fi

\begin{document}

\title{DeFormer: Integrating Transformers with Deformable Models for 3D Shape Abstraction from a Single Image}


\author{
Di Liu\textsuperscript{\rm 1},
Xiang Yu\textsuperscript{\rm 2},
Meng Ye\textsuperscript{\rm 1},
Qilong Zhangli\textsuperscript{\rm 1},
Zhuowei Li\textsuperscript{\rm 1},
Zhixing Zhang\textsuperscript{\rm 1},
Dimitris N. Metaxas\textsuperscript{\rm 1}
 \\
\textsuperscript{\rm 1}Rutgers University~
\textsuperscript{\rm 2}Amazon Prime Video\\
}

\maketitle
\ificcvfinal\thispagestyle{empty}\fi

\begin{abstract}
Accurate 3D shape abstraction from a single 2D image is a long-standing problem in computer vision and graphics. By leveraging a set of primitives to represent the target shape, recent methods have achieved promising results. However, these methods either use a relatively large number of primitives or lack geometric flexibility due to the limited expressibility of the primitives. In this paper, we propose a novel bi-channel Transformer architecture, integrated with parameterized deformable models, termed DeFormer, to simultaneously estimate the global and local deformations of primitives. In this way, DeFormer can abstract complex object shapes while using a small number of primitives which offer a broader geometry coverage and finer details. Then, we introduce a force-driven dynamic fitting and a cycle-consistent re-projection loss to optimize the primitive parameters. Extensive experiments on ShapeNet across various settings show that DeFormer achieves better reconstruction accuracy over the state-of-the-art, and visualizes with consistent semantic correspondences for improved interpretability.
\end{abstract}

\section{Introduction}
Accurate 3D shape abstraction with semantically meaningful parts is an active research field in computer vision for decades. It can be applied to many downstream tasks, such as shape reconstruction~\cite{mescheder2019occupancy,chen2019learning,park2019deepsdf,saito2019pifu,xu2019disn,michalkiewicz2019implicit,niemeyer2019occupancy,paschalidou2019superquadrics,liu2022deep}, object segmentation~\cite{hu2017deep,niu2018im2struct,li2023steering,ge2020automated,zou20173d,liu2020dispersion,liu2019dispersion,hatamizadeh2020end,liu2022transfusion,chang2022deeprecon,zhangli2022region,gao2022data,liu2021refined,liu2021label,eisenmann2022biomedical,he2023dealing,martin2023deep,gao2023training}, shape editing~\cite{yang2023movingparts,han2023improving} and re-targeting~\cite{deng2021deformed,halimi2019unsupervised,ye2021deeptag}. Due to the large success of deep neural networks (DNNs), a series of learning-based works~\cite{paschalidou2019superquadrics,paschalidou2020learning,tulsiani2017learning,paschalidou2021neural,deng2020cvxnet} propose to decompose an object shape into primitives and use the deformed primitives to represent the target shape. The primitive-based methods usually interpret a shape as a union of simple parts (\eg, cuboids, spheres, or superquadrics), offering interpretable abstraction of a shape target. 

\begin{figure}[t]
\centering
\includegraphics[width=1\linewidth]{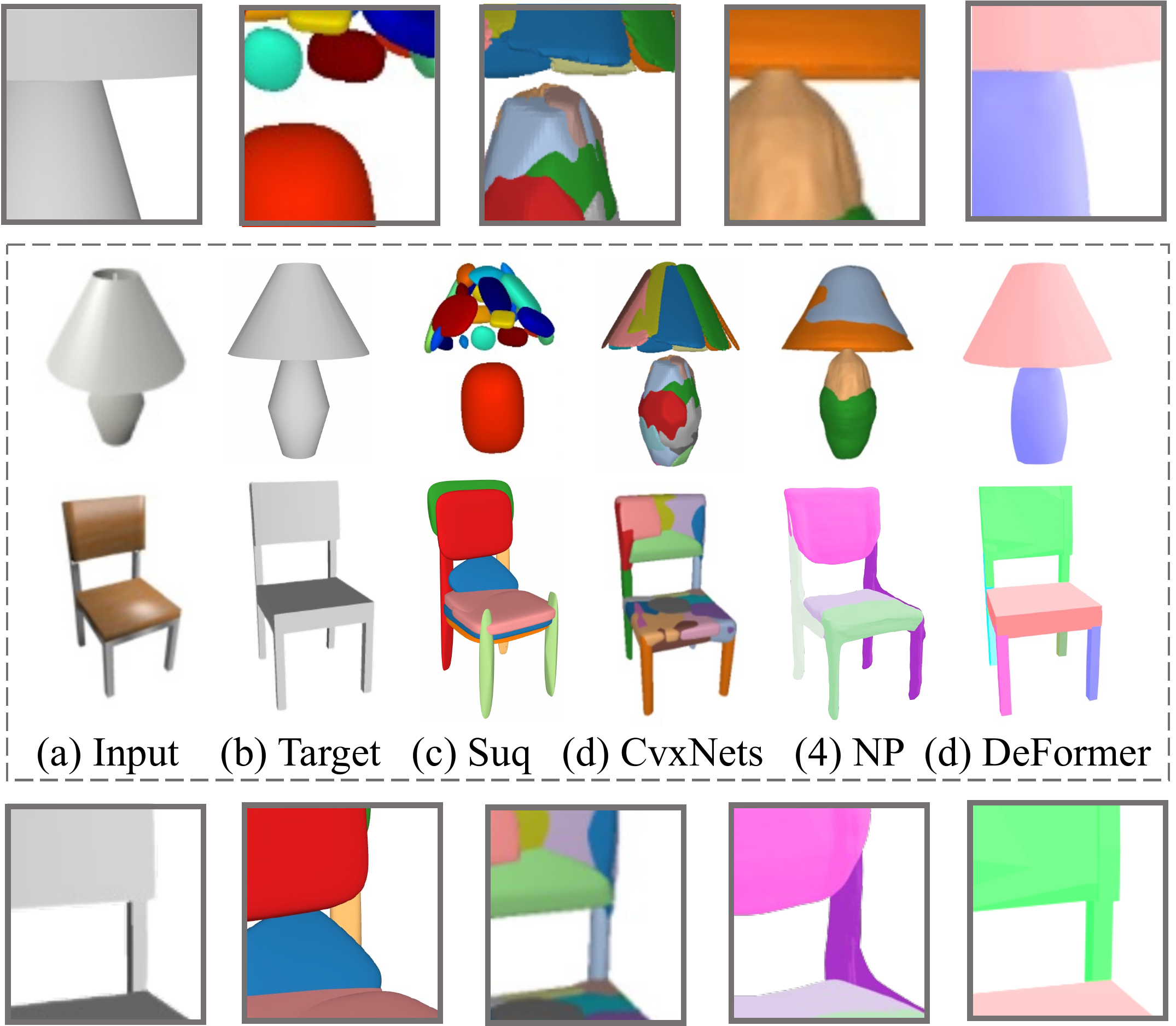}
\vspace{-6mm}
\caption{
DeFormer uses a small number of primitives to abstract a 3D shape from a 2D image with better accuracy and part correspondence. Taking ``lamp'' and ``chair'' as examples, we compare to Suq~\cite{paschalidou2019superquadrics}, CvxNets~\cite{deng2020cvxnet}, and Neural Parts (NP)~\cite{paschalidou2021neural} with $\sim$20, 25, and 5 primitives, respectively, while ours applies 2 primitives for lamps (1 for shade and 1 for body) and 6 primitives for chairs (4 for legs, 1 for seat and 1 for back).
}
\label{multi_prim_fig_cover}
\vspace{-6mm}
\end{figure}

To achieve high accuracy of shape reconstruction, existing methods require joint optimization of a number of primitives, which sometimes do not accurately correspond to the object parts and therefore limit the interpretability of the reconstructions~\cite{paschalidou2019superquadrics,deng2020cvxnet,paschalidou2021neural} (see Fig.~\ref{multi_prim_fig_cover}). To this end, using a small number of primitives to abstract complex shapes becomes a trend in recent research~\cite{paschalidou2021neural}. However, the dilemma lies in that using fewer primitives usually results in sub-optimal reconstruction accuracy due to their reduced representation power, while using more primitives lowers the interpretability and requires higher computational costs.

Physics-based deformable models (DMs)~\cite{metaxas2012physics,terzopoulos1991dynamic} are well known for their strong abstraction ability in shape representation, and have been successfully applied to various complex shape modeling applications. DMs leverage a physical modeling framework to predict global and local deformations of primitives, in which force-driven dynamic fitting across the data and the generalized latent space are used to jointly minimize the divergence between the deformed primitives and the target shapes. Although DMs can offer strong representation power for shape abstraction,  a main concern is that they require handcrafted parametric initialization for each specific shape abstraction, which limits their usage to general and automated shape modeling.

To address the aforementioned limitations, we propose a bi-channel Transformer combined with deformable models, termed {\it DeFormer}, to leverage the superior interpretability from DMs and overcome the parametric initialization limitation by taking advantage of the universal approximation capabilities~\cite{hornik1991approximation,hornik1989multilayer} of deep neural networks. Moreover, we leverage general superquadric primitives with global deformations as our primitive formulation, which offer a broader shape coverage and improve abstraction accuracy. To further enhance the shape coverage of the proposed DeFormer, we employ a diffeomorphic mapping that preserves shape topology to predict local deformations for finer details beyond the coverage of global deformations. 

To improve the primitive parameter optimization, we introduce ``external force" during training, to minimize the divergence between the deformed primitives and target shapes. This allows us to further use kinematic modeling for more flexible transformations across the data space, the generalized latent space, and the projected image space for improved robust training. To guarantee the training convergence, we leverage a cycle-consistent re-projection loss to achieve consistency between the reconstructed shapes, with the projected image and the original image as the input, respectively. Extensive experiments across several settings show that DeFormer outperforms the state-of-the-art (SOTA) with fewer primitives on the core thirteen shape categories of \textit{ShapeNet}. 

Our main contributions are summarized as follows:

\noindent $\bullet$ To the best of our knowledge, DeFormer is the first work that integrates Transformers with deformable models for accurate shape abstraction. We show that our novel learning formulation achieves better abstraction ability using a small number of primitives with a broader shape coverage.

\noindent $\bullet$ A force-driven dynamic fitting loss combined with a cycle-consistent re-projection regularization is introduced for effective and robust model training.

\noindent $\bullet$ Extensive experiments show that our method achieves better reconstruction accuracy and improved semantic consistency compared to the state-of-the-art.

\begin{figure*}[t]
  \centering
\includegraphics[width=1\linewidth]{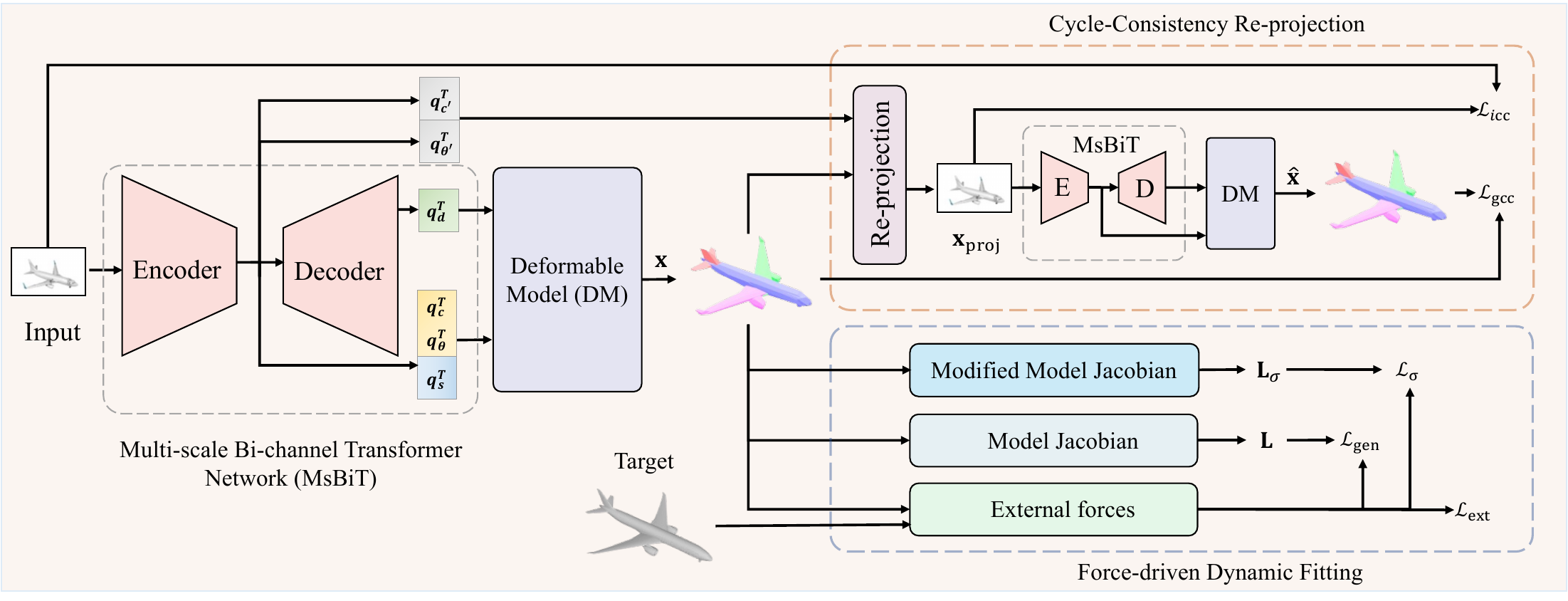}
\vspace{-2mm}
   \caption{DeFormer overview. Given an input image $\mathcal{X}$, a \textit{Multi-scale Bi-channel Transformer Network (MsBiT)} is proposed to hierarchically map $\mathcal{X}$ to a set of camera- and shape-related parameters that describe $P$ deformed primitives. The primitive parameters ${\textbf{q}}_c$, ${\textbf{q}}_\theta$, ${\textbf{q}}_s$, ${\textbf{q}}_d$, are passed through the deformable models to give a shape reconstruction $\textbf{x}$. To optimize the reconstruction, we employ a \textit{Force-driven Dynamic Fitting} module to minimize the forces applied to the primitives. To prevent overfitting to the training data, we propose a \textit{Cycle-Consistency Re-projection} loss for further regularization.}
   \vspace{-3mm}
   \label{framework}
\end{figure*}
\section{Related Work}

3D shape reconstruction can be categorized into several mainstreams. (1) Voxel-based methods~\cite{choy20163d,wu2016learning,gadelha20173d,jimenez2016unsupervised,chen2019learning} leverage voxels to capture 3D geometries. These methods usually require large memory and computation resources. Some methods reduce the memory cost~\cite{maturana2015voxnet,tatarchenko2017octree,hane2017hierarchical}, but the complexity of these frameworks increases significantly. (2) Point Cloud-based methods~\cite{fan2017point,qi2017pointnet++,achlioptas2018learning,jiang2018gal,thomas2019kpconv} require less computation but additional post-processing to address the lack of surface connectivity for mesh generation. (3) Mesh-based~\cite{kanazawa2018learning,wang2018pixel2mesh,pan2019deep,chen2019learning} can generate smooth shape surfaces, but they do not offer part-level decomposition of the shape. (4) Implicit \textcolor{black}{function}-based methods~\cite{mescheder2019occupancy,chen2019learning,park2019deepsdf,saito2019pifu,xu2019disn,michalkiewicz2019implicit,niemeyer2019occupancy} can also achieve high reconstruction accuracy of the shape, but they require heavy post-processing to obtain the final mesh. (5) Primitive-based methods~\cite{tulsiani2017learning,paschalidou2019superquadrics,paschalidou2020learning,hao2020dualsdf,deng2020cvxnet,paschalidou2021neural} represent object shapes by deforming a number of primitives, each of which is explicitly described by a set of \textcolor{black}{shape-related} parameters (\eg, scaling, squareness, tapering) whose properties are described in the following. 

\textbf{Primitive-based Shape Abstraction.}
Since our approach is primitive-based we thus focus on the most relevant primitive-based methods. 
Tulsiani \etal employ a union of cuboids to abstract object shapes~\cite{tulsiani2017learning}, while in \cite{paschalidou2019superquadrics,paschalidou2020learning}, Paschalidou \etal extend cuboids to superquadrics which provide extra geometric flexibility of the primitive. 
Other  \textcolor{black}{primitive} shapes such as spheres \cite{hao2020dualsdf} and convexes \cite{deng2020cvxnet} have also been investigated. The accuracy of these methods highly depends on the typically large number of primitives. Following this, Neural Parts~\cite{paschalidou2021neural} employ an Invertible Neural Network~\cite{ardizzone2018analyzing}
with reduced \textcolor{black}{number} of primitives to improve the performance. However, the primitives in Neural Parts do not often correspond to the object parts (especially those without clearly identified boundaries), thus resulting in reduced interpretability.
To address these limitations, we propose DeFormer with a small number of primitives \textcolor{black}{for more accurate shape} abstraction. We leverage deformable models to parameterize the primitives and guarantee the consistent correspondence between the primitives and the target shape, which significantly improves the abstraction ability. Another close work Pix2Mesh~\cite{wang2018pixel2mesh} is mesh-based, and applies a single template for shape deformation.
But it lacks explicit correspondence during deformation with the use of graph unpooling layers and may yield invalid mesh (\eg, self-intersecting mesh). In contrast, our deformation is diffeomorphic, which can preserve the topology of the primitive shape without breaking the connectivity in the mesh.

\textbf{Implicit Function-based Methods.} This set of methods mainly leverage implicit functions (\ie, level-sets) to directly estimate the signed distance function~\cite{chen2019bae,chen2019learning,park2019deepsdf,saito2019pifu,deng2021deformed,mescheder2019occupancy,genova2019learning,xu2019disn,li2021d2im,michalkiewicz2019implicit,niemeyer2019occupancy,zheng2021deep,Tertikas2023CVPR}. While they achieve high reconstruction accuracy, they usually need post-processing (\eg, marching cubes) to recover the shape surface. 
In contrast, primitive-based methods seek to decompose a target shape into parts and also decompose each part into explicit shape-related parameters (\eg, scaling, squareness, tapering, bending), which contribute to the understanding of primitive deformation. These shape-related parameters enable the explicit modeling for each shape part and provide semantic consistency among shapes.

\begin{figure}[t]
  \centering
\includegraphics[width=0.95\linewidth]{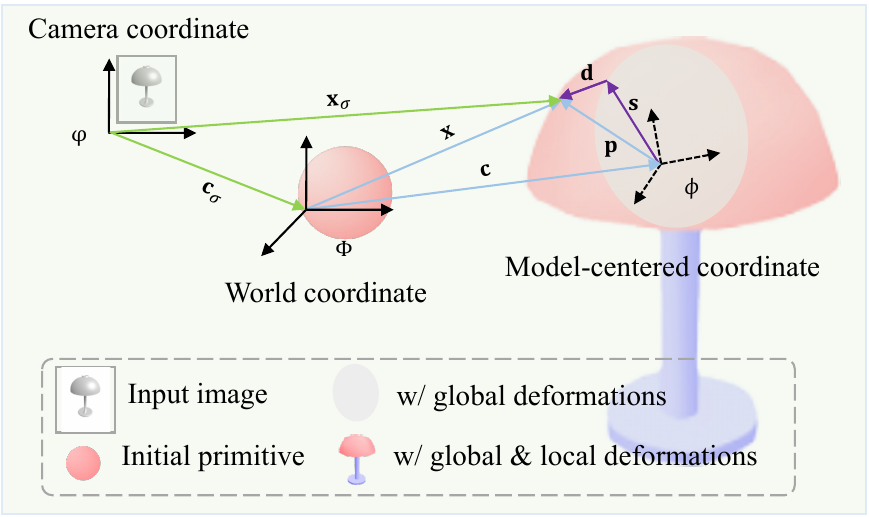}
   \caption{Generalized DeFormer geometry with perspective projection. It enables flexible transformations among the world coordinate $\Phi$, the model-centered coordinate $\phi$, and the camera coordinate $\varphi$.
   }
   \vspace{-5mm}
   \label{geometry}
\end{figure}
\section{Approach}
DeFormer (\cref{framework}) targets learning a set of primitives parameterized by DMs (\cref{method_prim}) given a single image as input. Each primitive is represented by a group of shape-related parameters ${\textbf{q}}$ which are estimated by the proposed \textit{Multi-scale Bi-channel Transformer Network (MsBiT)} (\cref{method_net}). A \textit{Force-driven Dynamic Fitting} module is introduced (\cref{method_loss}) to minimize the forces applied onto the primitives. To further improve the modeling accuracy, we propose a novel \textit{cycle-consistent re-projection} loss in \cref{method_proj} to regularize the estimated primitive deformations.

\subsection{Geometry and Primitive Formulation}
\label{method_prim}

\noindent \textbf{Canonical Geometry.} Following the physics-based deformable models \cite{terzopoulos1991dynamic,metaxas2012physics}, DeFormer assumes each individual primitive is a closed surface with a model-centered coordinate $\phi$.
As shown in \cref{geometry}, given a point $k$ on the primitive surface, its location $\textbf{x} = (x, y, z)$ \textit{w.r.t.} the world coordinate $\Phi$ is
\begin{equation}
    \textbf{x} = {\textbf{c}} + {\textbf{R}} \textbf{p} = {\textbf{c}} + {\textbf{R}} (\textbf{s} + \textbf{d}),
\label{x}
\end{equation}
where $\textbf{c}$ and ${\textbf{R}}$ represent the primitive translation and rotation \textit{w.r.t.} $\Phi$; $\textbf{p}$ denotes the relative position of the point $k$ on the primitive surface \textit{w.r.t.} $\phi$, which includes global deformation $\textbf{s}$ and local deformation $\textbf{d}$. 

\begin{figure*}[t]
  \centering
\includegraphics[width=1\linewidth]{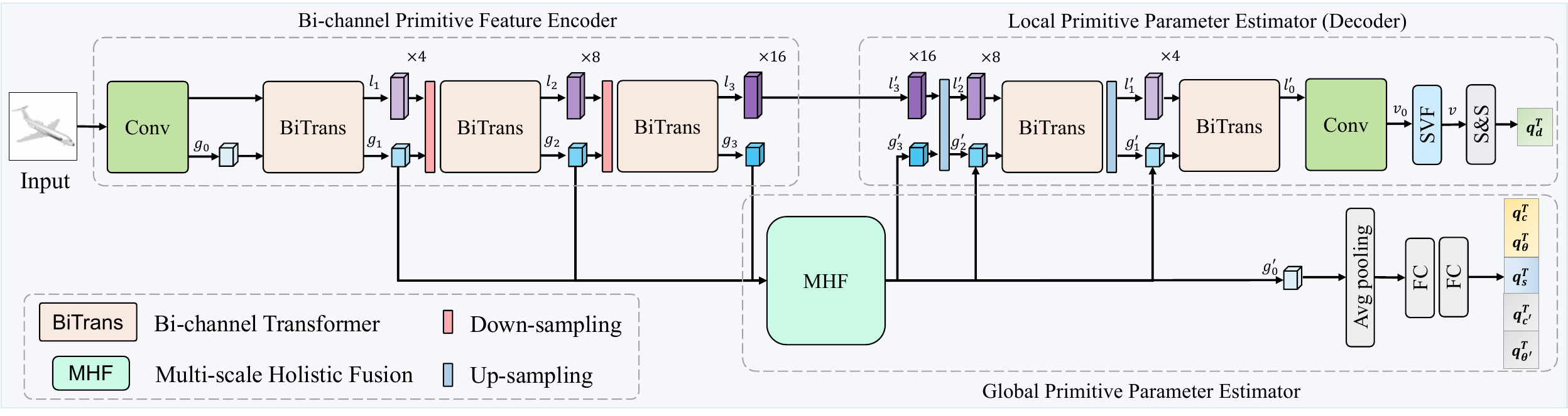}
\vspace{-5mm}
   \caption{The architecture of Multi-scale Bi-channel Transformer Network (MsBiT). Given an input image, MsBiT hierarchically predicts a set of primitive parameters. The \textit{Bi-channel Primitive Feature Encoder} first maps the input into two feature branches for global and local representations, $g_i$ and $l_i$, respectively. The encoded holistic maps $g_i$ are then passed through the \textit{Global Primitive Parameter Estimator} which outputs the final global shape-related parameters ${\textbf{q}}_c$, ${\textbf{q}}_\theta$, ${\textbf{q}}_{c'}$, ${\textbf{q}}_{\theta '}$, ${\textbf{q}}_s$. We employ the \textit{Local Primitive Parameter Estimator} to collect the encoded features and pass them through a diffeomorphic mapping for the final estimation of local deformation ${\textbf{q}}_d$.}
   \vspace{-4mm}
   \label{net}
\end{figure*}

\vspace{1mm}
\noindent \textbf{Generalized Geometry with Perspective Projection.} We seek to abstract the object shape in the world coordinate $\Phi$ given a single image $\mathcal{X}$, where $\mathcal{X}$ is in the camera reference frame $\varphi$ (See \cref{geometry}). Theoretically, the camera has a certain relative orientation corresponding to $\mathcal{X}$. To enable robust 3D shape abstraction that matches the 2D observation, we integrate the camera parameters into our current geometry for accurate camera pose estimation. Let ${\textbf{x}}_\sigma = (x_\sigma, y_\sigma, z_\sigma)$ be the location of the point $k$ on the primitive surface \textit{w.r.t.} the camera frame $\varphi$. Similar to \cref{x} we denote:
\begin{equation}
  \textbf{x}_\sigma = {\textbf{c}}_\sigma + {\textbf{R}}_\sigma \textbf{x},
\label{x_sigma}
\end{equation}
where ${\textbf{c}}_\sigma$ and ${\textbf{R}}_\sigma$ are the translation and rotation of the camera frame $\varphi$ \textit{w.r.t.} the world coordinate frame $\Phi$.

In summary, we decompose the transformations between the primitive and the target shape as camera translation $\textbf{c}_\sigma $, camera rotation $\textbf{R}_\sigma $, primitive translation $\textbf{c}$, primitive rotation $\textbf{R}$, global deformations $\textbf{s}$ and local deformations $\textbf{d}$. 

\vspace{1mm}
\noindent \textbf{Primitive Parameterization.}
We use superquadric surfaces with global and local deformations to represent the primitives due to their broad geometry coverage. We follow the original formulation of superquadrics in \cite{barr1981superquadrics,terzopoulos1991dynamic,liu2023deep} and employ a modified version by introducing global tapering and bending deformations as well as diffeomorphic local deformations. We provide a detailed formulation of our primitives in \textit{suppl. material}.

\subsection{Multi-scale Bi-channel Transformer Network}
\label{method_net}

\noindent \textbf{Bi-channel Transformer (BiTrans) Module.}  
Since the local receptive field of CNNs limits the modeling of global primitive features~\cite{vaswani2017attention,shaw2018self}, we employ Transformers to collect long-range dependencies for the prediction of the global primitive parameters. Geometrically, the primitive parameters related to the translation, rotation and global deformations have a holistic point of view to preserve the most salient primitive features, which makes the all-to-all attention in the Multi-head self-attention (MHSA)~\cite{vaswani2017attention} highly redundant. To address this,
we propose a Bi-channel Transformer (BiTrans) with two channels:
1) a low-dimensional feature map $g_i$ (blue) to preserve the holistic information, and 2) a conventional feature map $l_i$ (purple) to embed the local non-rigid information. 
The local deformation map $l_i$ is firstly projected to $Q/K/V$ with depth-wise separable convolution~\cite{chollet2017xception}. We employ 1 $\times$ 1 convolution to project $g_i$ with a much smaller size to $\overline Q/ \overline K/ \overline V$ to avoid any additional noise introduced in the depth-wise separable convolution padding. Due to the symmetry of the query and key dot product, we achieve the cross-attention map by transposing the dot product matrix to aggregate the global and local information of the primitive:
\begin{equation}
\begin{split}
(l_i^j, g_i^j) & = \text{BiTrans}(l_i^{j-1},g_i^{j-1})\\
&= (\text{softmax}(\frac{Q \overline{K} ^\top}{\sqrt{d}}) \overline{V}, \text{softmax}(\frac{\overline{Q} K ^\top}{\sqrt{d}})  V) ,   
\end{split}
\end{equation}
where $l_i^j$ and $g_i^j$ are the ``Bi-channel'' $j$-th layer outputs.

\vspace{1mm}
\noindent  \textbf{Global Primitive Parameter Estimator.}
We employ an estimator with an average pooling and two fully connected layers (\cref{net}) to map the embedded holistic features $g_0'$ to  global parameters ${\textbf{q}}_c$, ${\textbf{q}}_\theta$, ${\textbf{q}}_{c'}$, ${\textbf{q}}_{\theta '}$, ${\textbf{q}}_s$ corresponding to $\textbf{c}$, $\textbf{R}$, $\textbf{c}_\sigma$, $\textbf{R}_\sigma$, $\textbf{s}$, respectively.
\textcolor{black}{Specifically, ${\textbf{q}}_c = \textbf{c}$, ${\textbf{q}}_{c'} = \textbf{c}_\sigma$. ${\textbf{q}}_{\theta }$ and ${\textbf{q}}_{\theta '}$ are two four-dimensional quaternions related to $\textbf{R}$ and $\textbf{R}_\sigma$ defined in~\cite{terzopoulos1991dynamic}. ${\textbf{q}}_s = (a, \epsilon, t, b)$ determines the scaling $a$, squareness $\epsilon$, tapering $t$ and bending $b$ parameters of each primitives.}

\noindent \textbf{Local Primitive Parameter Estimator.}
To capture the finer shape details beyond the coverage of global deformations, we employ a diffeomorphic mapping to estimate the local non-rigid deformations $\textbf{q}_d = \textbf{d}$. Since the deformation with diffeomorphism is differentiable and invertible~\cite{arsigny2006log,dalca2019unsupervised}, it guarantees one-to-one mapping and preserves topology during the non-rigid deformations of the primitives.
Specifically, given the local features $l_0'$ from the MsBiT decoder, we first use a convolution stem to map $l_0'$ to a vector field $v_0$, and then map $v_0$ to a stationary velocity field (SVF) $v$ using a Gaussian smoothing layer. 
We follow~\cite{arsigny2006log,dalca2019unsupervised,dalca2018unsupervised,ashburner2007fast} and employ an Euler integration with a scaling and squaring layer (S\&S) to obtain the final local deformation $\textbf{q}_d$.

\subsection{\textcolor{black}{Force-driven Dynamic Fitting}}
\label{method_loss}

Similar to DMs, the primitives of DeFormer are able to dynamically deform to fit the target shape under the influence of external forces. Following the principle of virtual work\footnote{In mechanics, virtual work is the total work done by the applied forces of a mechanical system as it moves through a set of virtual displacements.}, we express the energy of the primitive as ${\mathcal E}_f  = \int {{f ^\top}d{\textbf{x}}} $. $f$ denotes the external force which measures how well the primitives are deformed to fit the target shape in data space (\textit{i.e.}, point-wise difference between the primitive surface and the target shape). When the primitive is far from the target, the force is large; vice versa. To optimize ${\mathcal E}_f$ we designate three specific loss terms as follows.

\noindent \textbf{External Model Loss ${{\mathcal L}_\text{ext}}$.}
We first employ the external model loss 
${{\mathcal L}_\text{ext}}$ to minimize the external forces applied to the $p$-th primitive, $f^p$, as: 
\begin{equation}
{{\mathcal L}_{{\text{ext}}}} = \frac{1}{P}\sum\limits_{p = 1}^{P}  {f^p}   = \frac{\gamma }{P}\sum\limits_{p = 1}^{P} {   {\mathcal D}({{{\mathcal M}_p,{\mathcal T}} }) },
\label{l_ext}
\end{equation}
where $\gamma$ is a constant modeling the strength of $f^p$
and $\mathcal{D}(\cdot)$ is the distance function that measures the difference between the points ${\textbf{x}^p_m}$ on the $p$-th deformed primitive $\mathcal M _p$ and the points $\tau _n$ on the target shape $\mathcal{T}$. Specifically, we employ a bi-directional Chamfer Distance (CD) for $\mathcal{D}(\cdot)$, denoted as:
\begin{equation}  
    \begin{split}
    {\mathcal D} ({{{\mathcal M}_p,{\mathcal T}} }) &= \frac{1}{\lvert {\mathcal{M}_p} \rvert}\sum\limits_{m = 1}^{\lvert {\mathcal{M}_p}\rvert} \min\limits_{\tau _n \in \mathcal{T}} {\Vert {\textbf{x}^p_m} - \tau _n \Vert}_2^2 \\
    &+ \frac{1}{\lvert {\mathcal{T}} \rvert}\sum\limits_{n = 1}^{\lvert {\mathcal{T}}\rvert}  \min\limits_{{\textbf{x}^p_m} \in {\mathcal{M}}_p} { \Vert \tau _n -{\textbf{x}^p_m}  \Vert}_2^2. 
   \label{df}
   \end{split}
\end{equation}

\noindent \textbf{Generalized Model Loss ${{\mathcal L}_\text{gen}}$.}
${{\mathcal L}_\text{ext}}$ can be viewed as a standard loss term for shape reconstruction, which, however, only controls the surface points of the predicted primitives with loose constraints. In addition, we seek to regularize the prediction by constraining each sub-transformation (\ie, primitive translation $\textbf{c}$, primitive rotation $\textbf{R}$, global $\textbf{s}$ and local deformations $\textbf{d}$) during dynamic fitting. Inspired by the kinematics of DMs~\cite{terzopoulos1991dynamic}, we achieve this by converting the forces computed in data space to the generalized forces in the generalized latent space.
Specifically, the kinematics are computed by $d {\textbf{x}} = {\textbf{L}} d {\textbf{q}}$, where $\textbf{L}$ is the Model Jacobian matrix that includes the Jacobians for $\textbf{c}$, $\textbf{R}$,  $\textbf{s}$, $\textbf{d}$~\cite{metaxas2012physics}. $\textbf{q}$ is the group of parameters controlling these sub-transformations. Then ${{\mathcal E}_f}$ is expressed as: 
\begin{equation}
{{\mathcal E}_f}  = \int {{f ^\top}d{\textbf{x}}  = \int {{ f ^\top}{{\textbf{L}}}d{\textbf{q}} } = \int {{f _q}d{\textbf{q}} }},
\label{L}
\end{equation}
where $f_q$ is the generalized force that measures the corresponding parameter-wise difference for each sub-transformation in the generalized latent space. It is based on the Model Jacobian ${\textbf L}  = [{{\textbf{I}} , \textbf{B}  , \textbf{R}   \textbf{J}  , \textbf{R}  }]$, where $\textbf{R}$ the rotation matrix, ${\textbf{B}} = {{\partial {\textbf{Rp}}}}/{\partial {\textbf{q}_\theta }}$ a rotation-related matrix, and $\textbf{J}$ the Jacobian matrix~\cite{terzopoulos1991dynamic}. Then $f _q$ is derived as:
\begin{equation}
\begin{split}
{f_q } = {{{f } ^\top}{{\textbf{L}} }} & = [{{{f } ^\top}}, {{{ f} ^\top}{\textbf{B}} }, {{{f} ^\top}{\textbf{R}} {{\textbf{J}} }}, {{{ f} ^\top}{\textbf{R}} }] \\
 & = {[f _{c} ^\top,f _{\theta}  ^\top,f _{s}  ^\top,f _{d} ^\top} ], 
\end{split}
\label{f_q}
\end{equation}
where $f _{c}$, $f _{\theta}$, $f _{s}$ and $f _{d}$ are the generalized force terms for the four sub-transformations $\textbf{c}$, $\textbf{R}$,  $\textbf{s}$ and $\textbf{d}$, respectively. This shows how the generalized forces $f_q $ are related to the external force $f^p$ using the Model Jacobian matrix $\textbf{L}$. 

Therefore, to regularize each sub-transformation during dynamic fitting, we employ a generalized model loss ${{\cal L}_\text{gen}}$ that minimizes $f_q$:
\begin{equation}
    \begin{split}
{{\cal L}_\text{gen}}
 & = \sum\limits_{p = 1}^{P} ({(f_{c}^p} {) ^\top} +  {(f_{\theta} ^p} {) ^\top} +  {(f_{s}^p} {) ^\top} +  {(f_{d}^p} {) ^\top})\\
  =& \sum\limits_{p = 1}^{P} ({{{(f^p) } ^\top}} + {{{ (f^p)} ^\top}{\textbf{B}} } + {{{(f^p)} ^\top}{\textbf{R}} {{\textbf{J}} }} + {{{ (f^p)} ^\top}{\textbf{R}} }).
    \end{split}
\end{equation}

\textcolor{black}{Note that our formulation of $f^p$ is a scalar approximation of the external force and not a vector. However, by minimizing each point-wise CD, we actually observe an approximated optimization result, in the sense of minimizing each point-wise force leading to the minimized joint force.}

\noindent \textbf{Image Model Loss ${{\mathcal L}_\sigma}$.}
In addition to jointly optimize the primitive surface points using ${{\cal L}_\text{ext}}$ and the sub-transformations for the four primitive deformations using ${{\cal L}_\text{gen}}$, we also seek to optimize the sub-transformations for the camera translation $\textbf{c}_\sigma$ and camera rotation $\textbf{R}_\sigma$. Similarly, we achieve this by converting $f$ to the forces in the projected image space according to the generalized geometry with perspective projection presented in \cref{method_prim}.
Specifically, Given a point on the primitive surface with location $\textbf{x}_\sigma  = (x_\sigma, y_\sigma, z_\sigma)$, its corresponding point on the projected image is expressed as ${\textbf{x}}_\text{proj} = (x_\text{proj}, y_\text{proj})$, where $x_\text{proj} = x_\sigma \mathcal{F} / z_\sigma$, $y_\text{proj} = y_\sigma \mathcal{F} / z_\sigma$ with $\mathcal{F}$ a constant to represent the focal length of the camera. By taking the time derivative, we obtain $d {\textbf{x}}_\text{proj} = \textbf{P} d {\textbf{x}}_\sigma$, where
\begin{equation}
    \textbf{P} =  
    \begin{bmatrix}
       \mathcal{F} / z_\sigma & 0                      & -x_\sigma \mathcal{F}/ {z^2_\sigma} \\
       0                      & \mathcal{F} / z_\sigma & -y_\sigma \mathcal{F}/ {z^2_\sigma} \\
    \end{bmatrix}.
\end{equation}
Given \cref{x_sigma}
and the kinematics $d {\textbf{x}} = {\textbf{L}} d {\textbf{q}}$~\cite{metaxas2012physics}, we obtain: 
\begin{equation}
    d{\textbf{x}}_\text{proj} = \textbf{P} d{\textbf{x}}_\sigma = \textbf{P}d({\textbf{c}}_\sigma + {\textbf{R}}_\sigma \textbf{x}) = \textbf{P} {\textbf{R}}_\sigma d {\textbf{x}}.
\label{dx_proj}
\end{equation}
The above \cref{dx_proj} allows us to modify the Model Jacobian matrix $\textbf{L}$ in DMs with ${\textbf{L}}_\sigma = {\textbf{P}} {\textbf{R}}_\sigma {\textbf{L}}$ for our generalized deformable model geometry with perspective projection, where ${\textbf{L}}_\sigma$ is named the Modified Model Jacobian matrix. By replacing the ${\textbf{L}}$ in \cref{f_q} with ${\textbf{L}}_\sigma$, we obtain ${f_\sigma } = {{f  ^\top}{\textbf{P}} {\textbf{R}}_\sigma {\textbf{L}}}$, which allows us to transform the external forces $f$ to the projected image forces $f_\sigma$. Similar to \cref{l_ext}, the projected image model loss ${{\mathcal L}_\sigma}$ computed using ${f_\sigma }$ is then derived as:
\begin{equation}
    {{\mathcal L}_\sigma} = \frac{1}{P}\sum\limits_{p = 1}^{P}  {f_\sigma ^p}   = \frac{1}{P}\sum\limits_{p = 1}^{P}  ({f ^p})^\top {\textbf{P}} {\textbf{R}}_\sigma {\textbf{L}}.
\end{equation}

By combining the illustrated losses related to the external, generalized, and projected image forces together, we summarize the dynamic fitting loss as:
\begin{equation}
    {\mathcal L}_{f} = {{\mathcal L}_\text{ext}} + {{\mathcal L}_\text{gen}} + {{\mathcal L}_\sigma}.
\label{force_loss}
\end{equation}

\begin{figure}[t]
\centering
\includegraphics[width=1\linewidth]{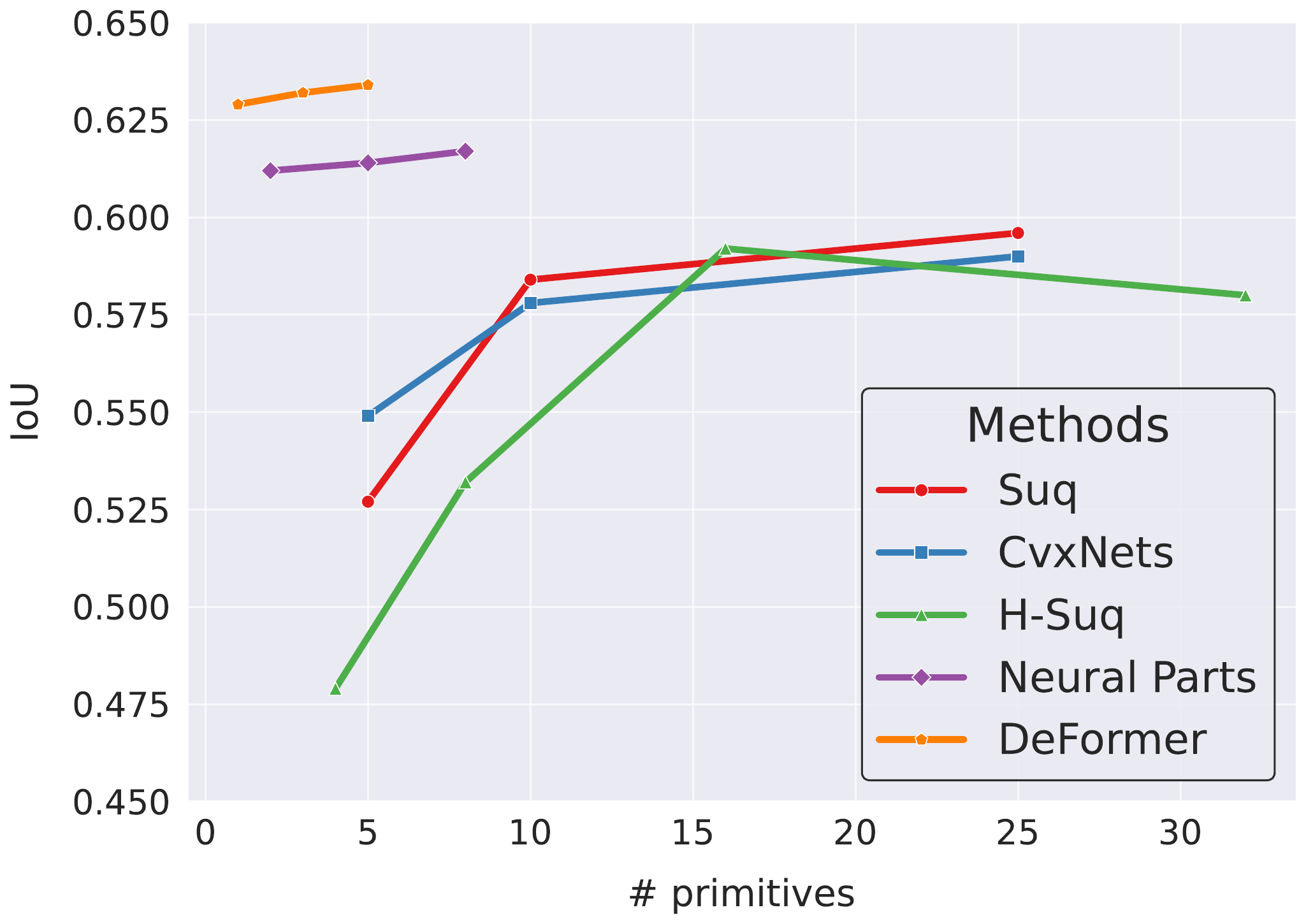}
\centering
\caption{Analysis of accuracy \textit{w.r.t.} the number of primitives used. \textcolor{black}{We focus on comparing to Neural Parts, with three data points, showing using a small number of primitives ($<$10) to achieve better reconstruction accuracy.}}
\label{prim_compare}
\end{figure}

\subsection{Cycle-Consistent Re-projection}
\label{method_proj}

To prevent network overfitting on the training data, we apply a differentiable re-projection module.
As shown in \cref{framework}, given the reconstructed primitive, we employ a differentiable renderer~\cite{liu2019soft} to re-project it onto the image domain using the predicted camera-related parameters ${\textbf{q}}_{c'}$, ${\textbf{q}}_{\theta '}$.
Then, by sending it to the network again, we expect DeFormer to have the same shape reconstruction as $\textbf{x}$. This process is formulated as a cycle-consistency regularization:
\begin{equation}
{{\mathcal L}_{{\text{gcc}}}} = \frac{1}{P}\sum\limits_{p = 1}^{P}  {\hat f^p}   = \frac{\hat \gamma}{\lvert P \rvert}\sum\limits_{p = 1}^{P} {   {\mathcal D}({{{\hat {\mathcal M}}_p,{\mathcal M}_p} }) },
\label{l_gcc}
\end{equation}
where ${\hat \gamma}$ is the strength of the pseudo external forces $\hat f^p$, and ${\hat {\mathcal M}}_p$ denotes the $p$-th re-reconstructed primitive given the projected image $\textbf{x}_\text{proj}$ as input.
If the above re-reconstruction is optimized, the projected image $\textbf{x}_\text{proj}$ should also match the original input $\mathcal{X}$. Therefore, we employ the image-level cycle-consistency loss $\mathcal{L}_\text{icc}$ to minimize the difference between $\textbf{x}_\text{proj}$ and $\mathcal{X}$:
\begin{equation}
{{\mathcal L}_{{\text{icc}}}} = \frac{1}{P}\sum\limits_{p = 1}^{P}  {\hat f^p_\sigma}   = \frac{\hat \gamma}{\lvert P \rvert}\sum\limits_{p = 1}^{P} \min {\Vert \textbf{x}_\text{proj}^{p} - \mathcal{X} \Vert}_2^2,
\label{l_icc}
\end{equation}
where $\hat f^p_\sigma$ is the pseudo image force and $\textbf{x}_\text{proj}^{p}$ is the image projected from the $p$-th primitive $\textbf{x}^p$.
Together with the dynamic fitting loss $\mathcal{L}_{f}$, we obtain the overall optimization objective as:
\begin{align}
    \mathcal{L} = \mathcal{L}_{f} + \mathcal{L}_\text{gcc} + \mathcal{L}_\text{icc}.
\label{total_loss}
\end{align}

\begin{figure}[t]

\centering
\includegraphics[width=1\linewidth]{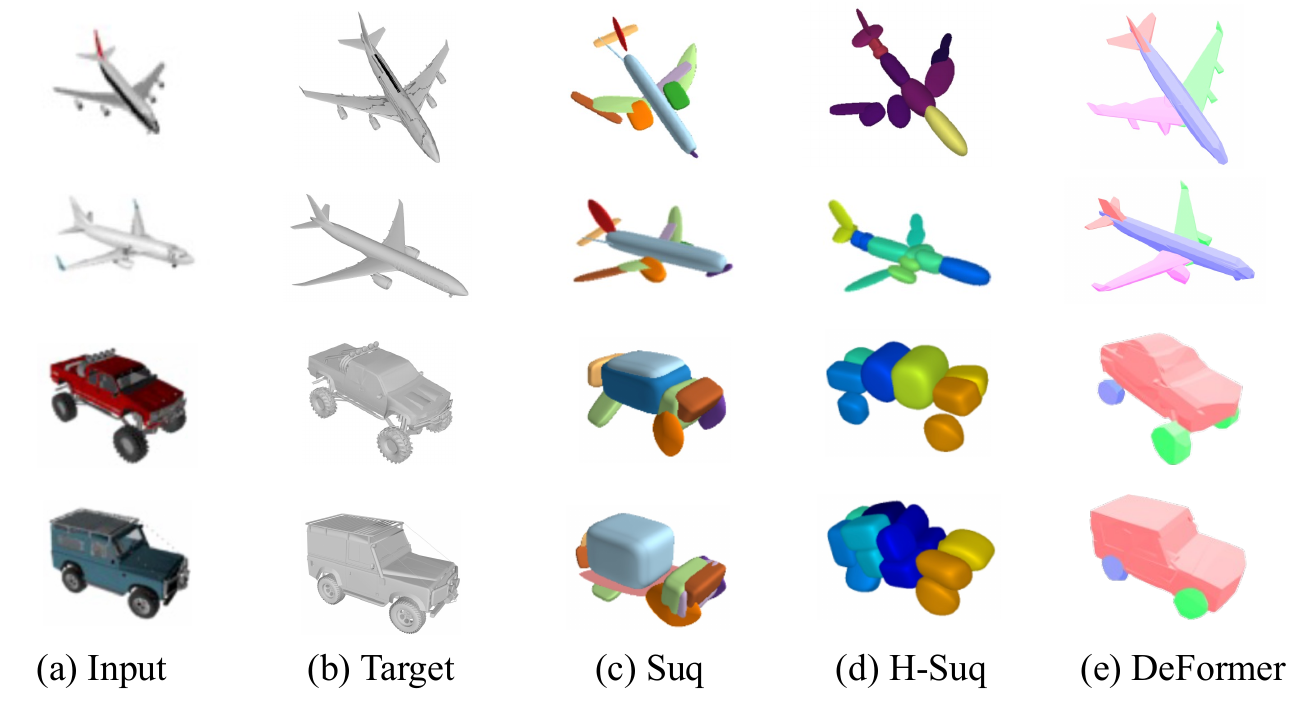}
\caption{Abstraction visualization compared to superquadrics-based methods, including Suq~\cite{paschalidou2019superquadrics} and H-Suq~\cite{paschalidou2020learning} with $\sim$20 primitives. In contrast, our model yields more accurate reconstructions with significantly fewer primitives (4 for airplanes and 3 for cars).}
\label{comp_suq_formulation}
\end{figure}

\begin{table*}[t]
\centering
\caption{Reconstruction results on the thirteen categories of \textit{ShapeNet}. We evaluate DeFormer (4) against P2M~\cite{wang2018pixel2mesh}, SIF~\cite{genova2019learning} (50), OccNet~\cite{mescheder2019occupancy}, Suq~\cite{paschalidou2019superquadrics} ($\leq 64$), CvxNets~\cite{deng2020cvxnet} (25), H-Suq~\cite{paschalidou2020learning} ($\leq 64$), and NP~\cite{paschalidou2021neural} (5). The Abs. Gain shows an absolute improvement to the second best. \textcolor{black}{Numbers in ($\cdot$) indicates the number of primitives used.}}

\label{Table3d}
\renewcommand\tabcolsep{4pt}
\resizebox{1\linewidth}{!}{
\begin{tabular}{l|ccc|ccccc|c|ccc|ccccc|c}
\toprule
\multirow{2}{*}{Category} & \multicolumn{9}{c|}{IoU ($\uparrow$)} & \multicolumn{9}{c}{Chamfer-$L_1$ ($ \downarrow $)}  \\ 
\cmidrule(lr){2-10}\cmidrule(lr){11-19}
       & P2M & SIF & OccNet  & Suq  & CvxNets& H-Suq   & NP      & DeFormer &  Abs. Gain   &P2M  &SIF & OccNet  & Suq   & CvxNets & H-Suq   & NP  & DeFormer & Abs. Gain \\ 
\midrule
airplane & 0.420 & 0.530 & 0.571 & 0.456    & 0.598    & 0.529    & 0.611     & \textbf{0.641}  & 3.0$\%$ & 0.187 & 0.167 & 0.147 & 0.122    & 0.093    & 0.175    & 0.089   & \textbf{0.072} & 0.017 \\
bench    & 0.323 & 0.333 & 0.485 & 0.202    & 0.461    & 0.437    & 0.502     & \textbf{0.528}  & 2.6$\%$ & 0.201 & 0.261 & 0.155 & 0.114    & 0.133    & 0.153    & 0.108   & \textbf{0.087} & 0.021 \\
cabinet  & 0.664 & 0.648 & 0.733 & 0.110    & 0.709    & 0.658    & 0.681     & \textbf{0.717}   & 0.4$\%$ & 0.196 & 0.233 & 0.167 & 0.087    & 0.102    & 0.087    & 0.083   & \textbf{0.074} & 0.009 \\
car      & 0.552 & 0.657 & 0.737 & 0.650    & 0.675    & 0.702    & 0.719     & \textbf{0.729} & 1.0$\%$ & 0.180 & 0.161 & 0.159 & 0.117    & 0.103    & 0.141    & 0.127   & \textbf{0.093} & 0.010 \\
chair    & 0.396 & 0.389 & 0.501 & 0.176    & 0.491    & 0.526    & 0.532     & \textbf{0.551}  & 1.9$\%$ & 0.265 & 0.380 & 0.228 & 0.138    & 0.337    & 0.114    & 0.107   & \textbf{0.089} & 0.018 \\
display  & 0.490 & 0.491 & 0.471 & 0.200    & 0.576    & 0.633    & 0.646     & \textbf{0.653}   & 0.7$\%$ & 0.239 & 0.401 & 0.278 & 0.106    & 0.223    & 0.137    & 0.098   & \textbf{0.087} & 0.011 \\
lamp     & 0.323 & 0.260 & 0.371 & 0.189    & 0.311    & 0.441    & 0.402     & \textbf{0.442}  & \textbf{4.0}$\%$ & 0.308 & 1.096 & 0.479 & 0.189    & 0.795    & 0.169    & 0.153   & \textbf{0.141} & 0.012 \\
speaker  & 0.599 & 0.577 & 0.647 & 0.136    & 0.620    & 0.660    & 0.693     & \textbf{0.715}  & 2.2$\%$ & 0.285 & 0.554 & 0.300 & 0.132    & 0.462    & 0.108    & 0.128   & \textbf{0.092}& 0.016 \\
rifle    & 0.402 & 0.463 & 0.474 & 0.519    & 0.515    & 0.435    & 0.537     & \textbf{0.540}   & 0.3$\%$ & 0.164 & 0.193 & 0.141 & 0.127    & 0.106    & 0.203    & 0.189   & \textbf{0.089} & 0.017 \\
sofa     & 0.613 & 0.606 & 0.680 & 0.122    & 0.677    & 0.693    & 0.712     & \textbf{0.729}  & 1.7$\%$ & 0.212 & 0.272 & 0.194 & 0.106    & 0.164    & 0.128    & 0.107   & \textbf{0.088} & 0.018 \\
table    & 0.395 & 0.372 & 0.506 & 0.180    & 0.473    & 0.491    & 0.531     & \textbf{0.546}  & 1.5$\%$ & 0.218 & 0.454 & 0.189 & 0.110    & 0.358    & 0.122    & 0.102   & \textbf{0.081} & 0.021 \\
phone    & 0.661 & 0.658 & 0.720 & 0.185    & 0.719    & 0.770    & 0.810     & \textbf{0.822}  & 1.2$\%$ & 0.149 & 0.159 & 0.140 & 0.112    & 0.083    & 0.149    & 0.076   & \textbf{0.064} & 0.012 \\
vessel   & 0.397 & 0.502 & 0.530 & 0.471    & 0.552    & 0.570    & 0.605     & \textbf{0.630}   & 2.5$\%$ & 0.212 & 0.208 & 0.218 & 0.125    & 0.173    & 0.178    & 0.119   & \textbf{0.096} & 0.023 \\
\midrule
Average  & 0.480 & 0.499 & 0.571 & 0.277    & 0.567    & 0.580    & 0.614     & \textbf{0.634} & 1.8$\%$ & 0.216 & 0.349 & 0.215 & 0.122    & 0.245    & 0.143    & 0.114   & \textbf{0.089} & \textbf{0.025}\\
\bottomrule
\end{tabular}} 
\vspace{-2mm}
\end{table*}

\section{Experiments}
\label{experiments}

\noindent \textbf{Datasets.} 
\textcolor{black}{We evaluate on \textit{ShapeNet}~\cite{chang2015shapenet}, a richly-annotated, large-scale dataset of 3D shapes. 
A subset of \textit{ShapeNet} including 50k models and 13 major categories are used in our experiments. We use the rendered views from 3D-R2N2~\cite{choy20163d}, and their training and testing split setting, which was the seminal work in the literature, and the setting has been utilized by most of the following-up papers.}

\noindent \textbf{Baselines.} Since DeFormer lies in the primitive-based mainstream with explicit representation, we mainly compare it to primitive-based methods, \ie, Suq~\cite{paschalidou2019superquadrics} and H-Suq~\cite{paschalidou2020learning} that both use superquadrics, CvxNets~\cite{deng2020cvxnet} using convexes, and NP~\cite{paschalidou2021neural} using spheres.
For non-primitive methods, we compare to SIF~\cite{genova2019learning}, P2M~\cite{wang2018pixel2mesh}, and OccNet~\cite{mescheder2019occupancy} which, however, lack explicit understanding of part correspondence.

\subsection{Implementation Details}
Throughout the training, Adam \cite{kingma2014adam} is employed for optimization and the learning rate is initialized as $10^{-4}$. We use a batch size of 32 and train the model for 300 epochs. All experiments are implemented with PyTorch and run on a Linux system with eight Nvidia A100 GPUs. Assuming input image size $H\times W$ and $d$ the token dimension, compared to CNNs complexity $\mathcal{O}(k^{2}HWd^{2})$ with convolution kernel size $k$, our BiTrans complexity is $\mathcal{O}(4HWd^{2}+2(HW)^{2}d)$, which achieves the same order of complexity as CNNs.

Similar to \cite{deng2020cvxnet,paschalidou2021neural,paschalidou2019superquadrics}, for each shape category with a certain number of primitives used, we train a separate model. We draw 2k random sample points from the surface of each target mesh as ground truth, and we sample 1k points from each generated primitive for shape reconstruction. During the evaluation, we uniformly sample 100k points on the target/predicted mesh to compute the volumetric Intersection over Union (IoU) and the Chamfer-$L1$ distance (CD). We empirically set the weights for the dynamic fitting loss in \cref{force_loss} as 0.5, 0.3, and 0.2, respectively. Similarly, we set the balance factors for the joint loss in \cref{total_loss} as 0.6, 0.2, and 0.2, respectively, for best performance. Ablation study for losses is provided in \cref{ab_tab_loss}. For the estimation of local deformations, we follow \cite{arsigny2006log,dalca2019unsupervised,dalca2018unsupervised,ashburner2007fast} and use $T=7$ scaling and squaring steps.

\begin{figure}[t]
\centering
\includegraphics[width=1\linewidth]{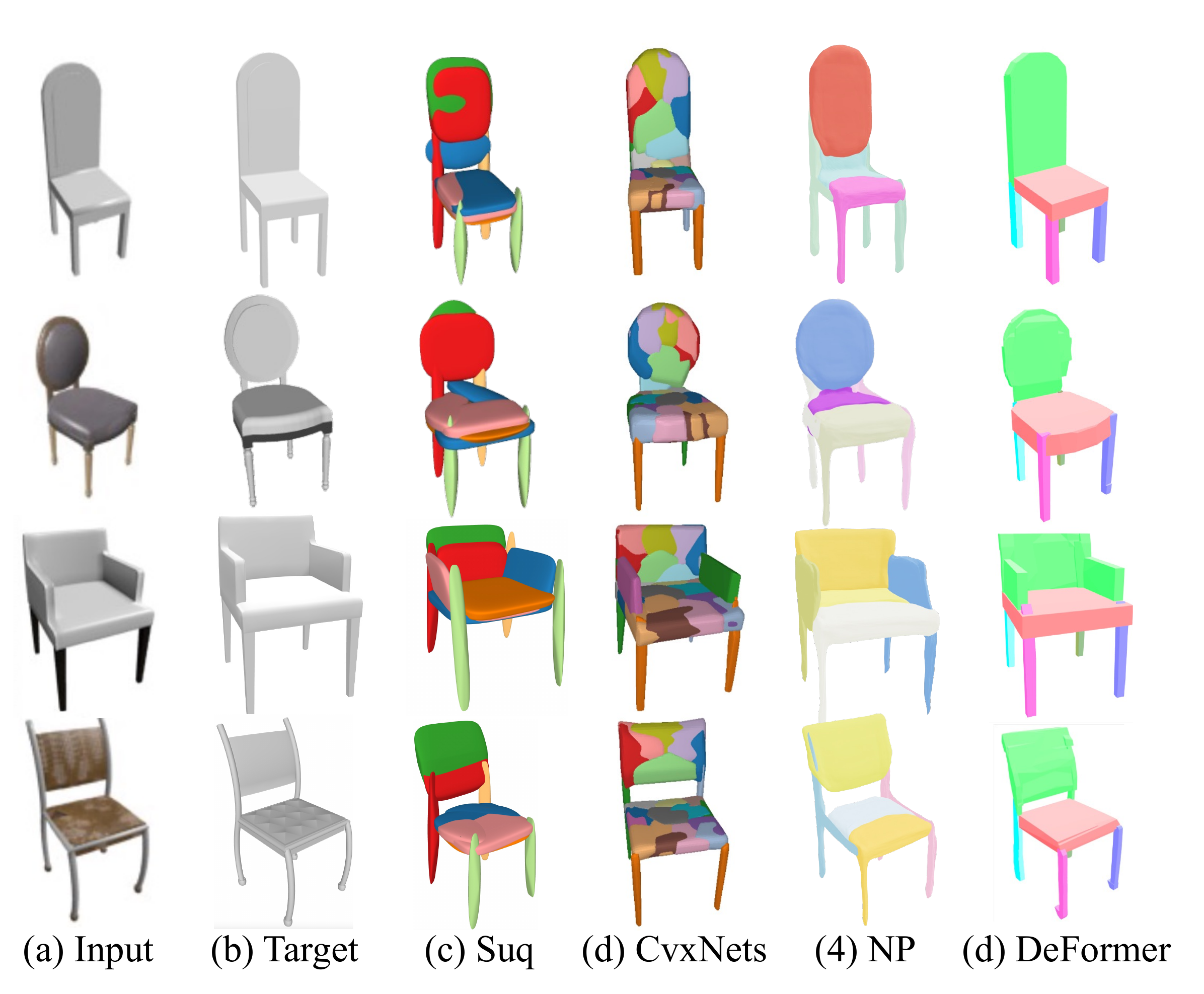}
\vspace{-6mm}
\caption{
Abstraction visualization on chairs compared to primitive-based methods, including Suq~\cite{paschalidou2019superquadrics}, CvxNets~\cite{deng2020cvxnet}, NP~\cite{paschalidou2021neural} with $\sim$20, 25 and 5 primitives, respectively. Ours applies 6 primitives (4 legs, 1 seat, and 1 back) and achieves better part consistency.}
\label{multi_prim_fig}
\vspace{-2mm}
\end{figure}

\begin{figure} [t]
\begin{center}
\vspace{10pt}
\includegraphics[width=1\linewidth]{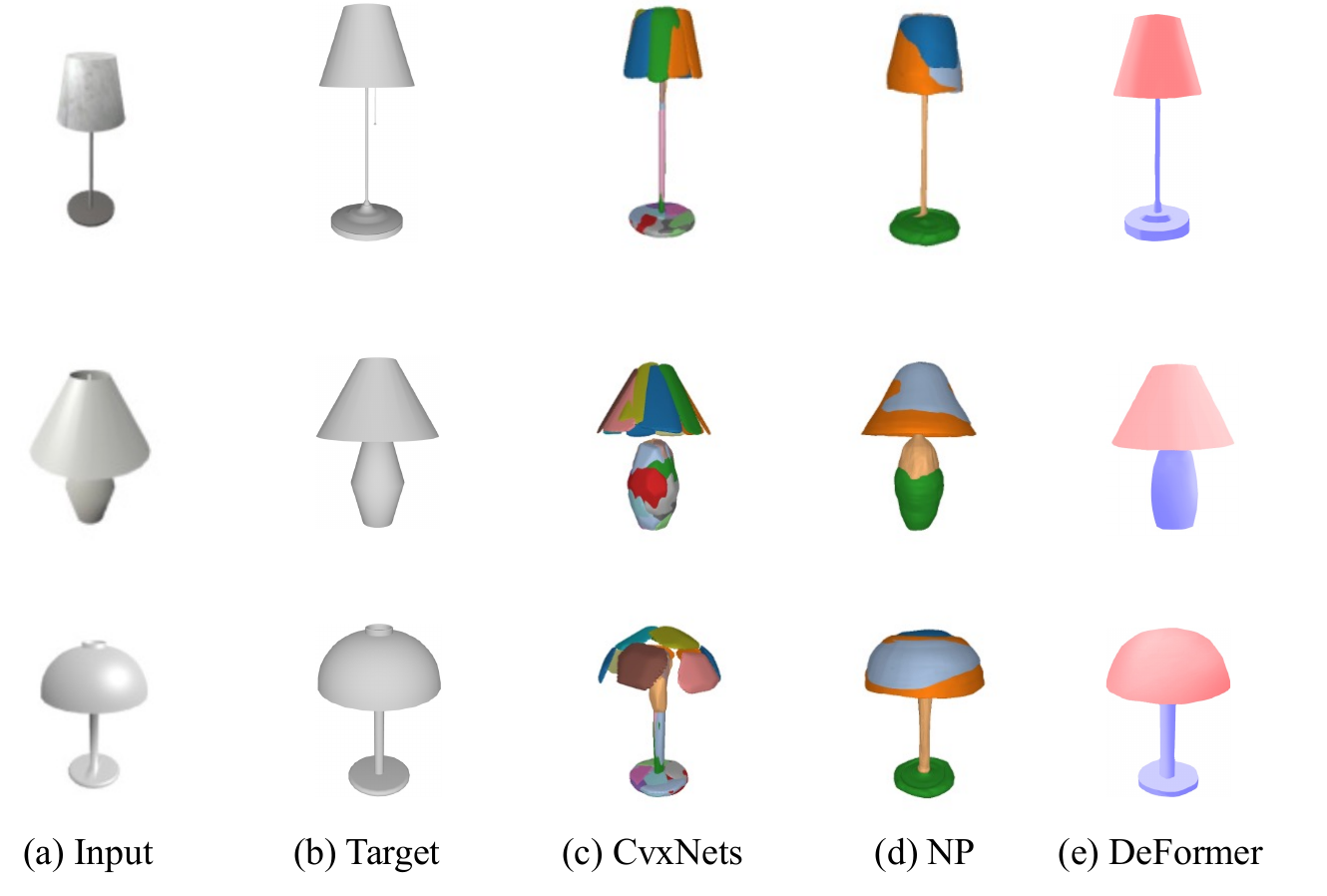}
\end{center}
\vspace{-5pt}
\caption{Abstraction visualization on lamps compared to SOTA primitive-based methods, including CvxNets~\cite{deng2020cvxnet} and NP~\cite{paschalidou2021neural} with 25 and 5 primitives, respectively. Ours applies 2 primitives (1 head and 1 base) and achieves better part consistency.}
\label{supp_lamp}
\end{figure}

\begin{figure*}[t]
  \begin{center}
\includegraphics[width=1\linewidth]{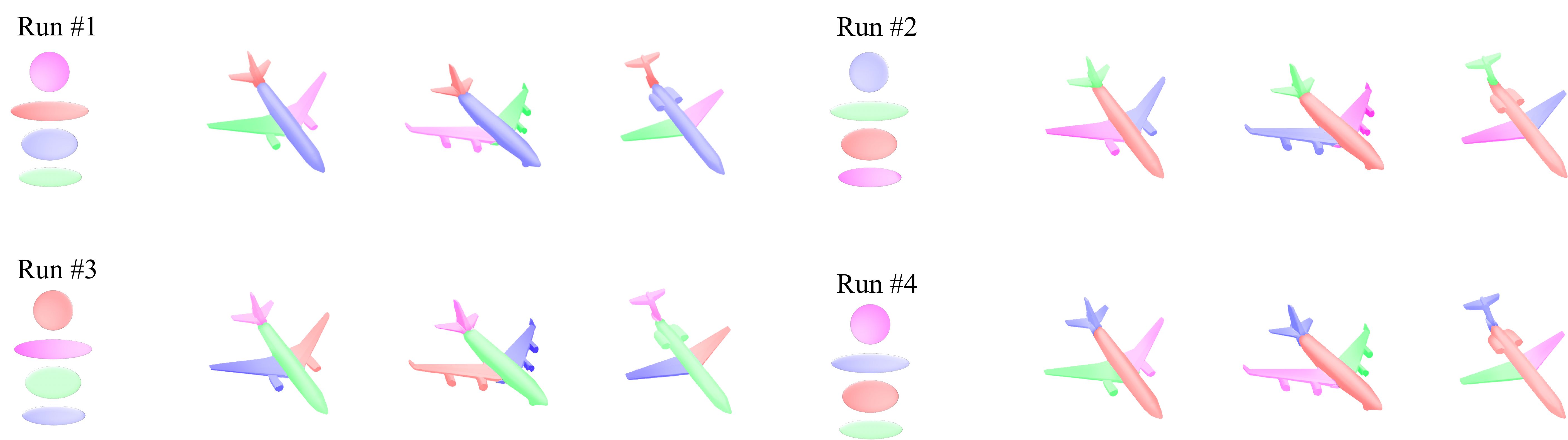}
  \end{center}
  \caption{\textcolor{black}{
  Illustration of semantic consistency. We set four different random seeds. For each seed, we observe consistent part correspondence (\eg, left-wing, tail, body and right-wing denoted as the same color) across the three ``airplane'' instances.
  }
  }
\label{ab_ini}
  \vspace{-8pt}
\end{figure*}

\subsection{Representation Power}
We first report the results of reconstruction accuracy \textit{w.r.t.} the number of primitives $P$ in \cref{prim_compare}. Our method shows consistently better IoU regardless of the number of primitives used.
We further see that the \textcolor{black}{reconstruction curve of DeFormer} saturates fast when the number of primitives increases. This is due to the broad geometric coverage of the proposed primitive formulation where a small number of primitives are sufficient for the optimal \textcolor{black}{shape abstraction}.
Moreover, to qualitatively demonstrate the representation superiority of our primitive formulation, we compare to Suq~\cite{paschalidou2019superquadrics} and H-Suq~\cite{paschalidou2020learning} with $\sim$ 20 primitives, which also use superquadrics in \cref{comp_suq_formulation}. We train DeFormer with fewer primitives ( 4 for airplanes and 3 for cars) and obtain better reconstruction accuracy and semantic consistency. 

\subsection{Reconstruction Accuracy}

We quantitatively evaluate the reconstruction performance against a number of SOTAs in \cref{Table3d}. Following their settings we train Suq~\cite{paschalidou2019superquadrics} and H-Suq~\cite{paschalidou2020learning} with a maximum of 64 primitives. For CvxNets~\cite{deng2020cvxnet} and SIF~\cite{genova2019learning} we report results with 25 primitives and 50 elements, respectively. For NP~\cite{paschalidou2021neural} and DeFormer, we use 5 and 4 primitives, respectively.
Note that P2M~\cite{wang2018pixel2mesh} and the implicit function-based methods OccNet~\cite{mescheder2019occupancy} and SIF~\cite{genova2019learning} are not directly comparable with the primitive-based methods, due to their lack of shape abstraction ability.  
Nevertheless, we observe from \cref{Table3d} that DeFormer outperforms all the SOTA results with on average $1.8\%$ IoU accuracy improvement and $2.5\%$ less Chamfer-$L_{1}$ distance.
We provide a qualitative comparison in \cref{multi_prim_fig} and \cref{supp_lamp}.

\subsection{Ablation Study}

\noindent \textbf{Semantic Consistency.} 
We investigate the ability of DeFormer to decompose 3D shapes into semantically consistent parts using different primitive initializations. Specifically, we train with four different random seeds on the airplane category and observe in Fig.~\ref{ab_ini} that the reconstructions preserve similar semantic parts for each seed.

\noindent \textbf{Loss Components.} 
In \cref{ab_tab_loss} using the ``leave-one-out'' way, each of the loss terms is highlighted and demonstrated to be a uniquely effective component within our overall loss term. Another observation is that training without $\mathcal{L}_\text{ext}$ results in a severe performance drop. The cycle-consistency losses $\mathcal{L}_\text{gcc}$ and $\mathcal{L}_\text{icc}$ provide key self-supervision for \textcolor{black}{unreasonable reconstruction} correction.

\begin{table}[t]
\centering
\renewcommand\tabcolsep{8pt}
\renewcommand{\arraystretch}{1}
\resizebox{0.47\textwidth}{!}{
\begin{tabular}{ccccc|ccc}
\toprule
 \multicolumn{5}{c|}{Settings} & \multicolumn{3}{c}{IoU ($\uparrow$) }  \\
\midrule
   ${{\cal L}_\text{ext}}$&  ${{\cal L}_\text{gen}}$ & ${{\cal L}_\sigma}$ & ${{\cal L}_\text{gcc}}$ & ${{\cal L}_\text{icc}}$ & car  & airplane & chair \\ 
\midrule

 \xmark & \cmark  & \cmark  & \cmark & \cmark & 0.718  & 0.633 & 0.547 \\

\cmark  &  \xmark & \cmark  & \cmark & \cmark &  0.723 & 0.641 & 0.550   \\

\midrule
\cmark & \cmark  & \xmark & \cmark &   \cmark & 0.729  & 0.648 & 0.556 \\

\cmark  & \cmark  & \cmark  & \xmark & \cmark & 0.731  & 0.647 & 0.552  \\

\cmark  & \cmark  & \cmark  & \cmark & \xmark & 0.733  & 0.652 & 0.559 \\

\midrule

\cmark  &  \cmark & \cmark  & \xmark & \xmark &  0.721  & 0.638 & 0.545 \\

\cmark  & \cmark  & \cmark  & \cmark & \cmark & \textbf{0.729}  & \textbf{0.641}  & \textbf{0.551}  \\

\bottomrule
\end{tabular}} 
\captionof{table}{Ablation studies on loss terms.  We report the average IoU on the major three categories of \textit{ShapeNet}.}
\label{ab_tab_loss}
\end{table}

\section{Conclusion}
We propose a novel bi-channel Transformer integrated with deformable models, termed DeFormer, to jointly predict global and local deformations for 3D shape abstraction. DeFormer achieves improved semantic correspondences thanks to the diffeomorphic mapping for shape estimation. Moreover, we leverage the force-driven dynamic fitting and the cycle-consistent re-projection loss to effectively optimize the shape parameters. Extensive experiments demonstrate our method achieves superior reconstruction performance and semantic consistency. Future work will consider more primitive formulations and global deformations for more general shape abstraction scenarios.

\subsection*{Acknowledgments}
This research has been partially funded by research grants to D. Metaxas through NSF: IUCRC CARTA 1747778, 2235405, 2212301, 1951890, 2003874, and NIH-5R01HL127661.

{\small
\bibliographystyle{ieee_fullname}
\bibliography{egpaper_final}
}


\end{document}